\PassOptionsToPackage{table}{xcolor}
\documentclass[10pt,twocolumn,letterpaper]{article}

\usepackage{cvpr}              

\definecolor{cvprblue}{rgb}{0.21,0.49,0.74}
\usepackage[pagebackref,breaklinks,colorlinks,allcolors=cvprblue]{hyperref}
\usepackage{enumitem}
\usepackage{multirow}
\usepackage{makecell}
\usepackage[table]{xcolor}
\usepackage{amssymb}
\usepackage{marvosym}
\def\paperID{13461} 
\def\confName{CVPR}
\def\confYear{2025}

\title{Distilled Prompt Learning for Incomplete Multimodal Survival Prediction}

\author{Yingxue Xu\textsuperscript{1} \quad Fengtao Zhou\textsuperscript{1} \quad Chenyu Zhao\textsuperscript{1} \quad Yihui Wang\textsuperscript{1} \quad Can Yang\textsuperscript{2} \quad Hao Chen\textsuperscript{1\Letter}\\
\textsuperscript{1}Department of Computer Science and Engineering, \textsuperscript{2}Department of Mathematics\\
The Hong Kong University of Science and Technology\\
{\tt\small \{yxueb,fzhouaf,cyzhao,ywangrm\}@connect.ust.hk, \{macyang,jhc\}@ust.hk}
}

\begin{document}
\maketitle
\begin{abstract}
The integration of multimodal data including pathology images and gene profiles is widely applied in precise survival prediction. Despite recent advances in multimodal survival models, collecting complete modalities for multimodal fusion still poses a significant challenge, hindering their application in clinical settings. Current approaches tackling incomplete modalities often fall short, as they typically compensate for only a limited part of the knowledge of missing modalities. To address this issue, we propose a Distilled Prompt Learning framework (DisPro) to utilize the strong robustness of Large Language Models (LLMs) to missing modalities, which employs two-stage prompting for compensation of comprehensive information for missing modalities. In the first stage, Unimodal Prompting (UniPro) distills the knowledge distribution of each modality, preparing for supplementing modality-specific knowledge of the missing modality in the subsequent stage. In the second stage, Multimodal Prompting (MultiPro) leverages available modalities as prompts for LLMs to infer the missing modality, which provides modality-common information. Simultaneously, the unimodal knowledge acquired in the first stage is injected into multimodal inference to compensate for the modality-specific knowledge of the missing modality. Extensive experiments covering various missing scenarios demonstrated the superiority of the proposed method. The code is available at \url{https://github.com/Innse/DisPro}.
\end{abstract}
\section{Introduction}
\label{sec:intro}
Multimodal Survival Prediction typically incorporates pathology slides and genomic data complementary to each other for precise prognosis analysis. Pathology data can provide qualitative morphological insights, while quantitative molecular information can be obtained from genomic data~\cite{chen2021multimodal,xu2023multimodal,xiong2024mome}. Despite progress~\cite{chen2021multimodal,xu2023multimodal,zhou2023cross} in multimodal survival analysis, these models typically require complete modalities data for reliable performance. However, the expectation of having full access to complete modalities for integration is not always achievable, especially in clinical scenarios, due to various factors such as the cost of data gathering and privacy issues. For example, acquiring genomic data still comes at a high cost due to the technology and infrastructure required for gene sequencing, especially in underdeveloped areas~\cite{zhou2024multimodal}. Inevitably, it is challenging to adapt previously established multimodal survival models~\cite{zhangprototypical,song2024multimodal} to real-world clinical settings. As a result, building a multimodal survival model that is robust to incomplete modalities data is an urgent issue for clinical practice.
\begin{figure}
    \centering
    \includegraphics[scale=0.495]{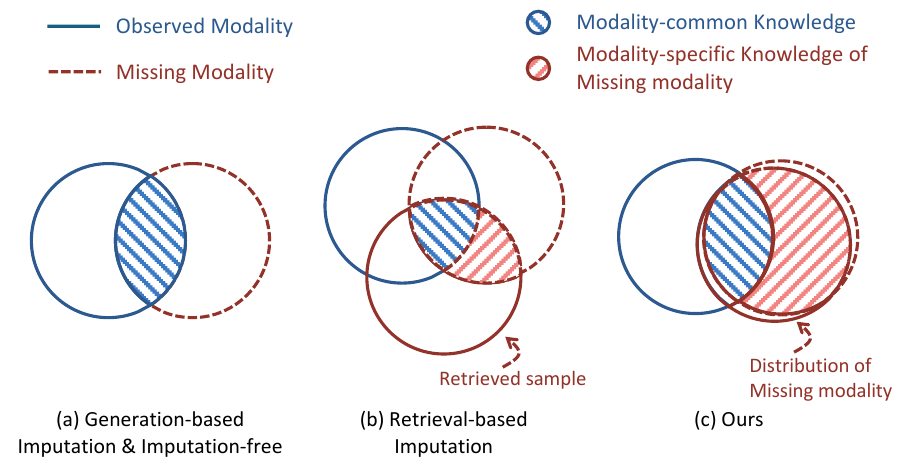}
    \vspace{-0.2cm}
    \caption{Insights for existing incomplete multimodal learning and comparison to the proposed method. (a) Generation-based Imputation and Imputation-free approaches, (b) Retrieved-based Imputation and (c) Ours.}
    \vspace{-0.6cm}
    \label{fig:intro}
\end{figure}

Currently, mainstream approaches addressing incomplete modalities issues include imputation-based and imputation-free methods~\cite{ding2021rfnet,wu2024multimodal}. The former approaches aim to impute the missing modality based on observed modalities, primarily achieved in two ways including generation~\cite{chartsias2017multimodal, meng2024multi,sharma2019missing} and retrieval~\cite{chen2020hgmf,zhang2022m3care}. In terms of generation-based imputation, generative models are typically employed for synthesizing features or raw data of missing modalities from available modalities. As shown in Fig.~\ref{fig:intro} (a), these methods usually can only impute modality-common information, as the modality-specific information unique to the missing modality cannot be generated out of thin air. Moreover, the goal of retrieval-based imputation methods is to fill in missing modalities by seeking the most similar sample of the missing modality from the training set. This way tends to compensate only a limited part of modality-common and modality-specific information, as shown in Fig.~\ref{fig:intro} (b). The reason is that an individual sample exhibits considerable randomness and is challenging to fully capture the unique knowledge inherent in the missing modality.

Unlike imputation-based approaches, imputation-free approaches~\cite{ding2021rfnet,wu2024multimodal} primarily focus on how to minimize performance degradation resulting from missing modalities. These approaches make great efforts to learn robust multimodal representations that are less sensitive to missing modalities by capturing modality-invariant information. Recently, with the rise of Large Language Models (LLMs)~\cite{biswas2023role,wu2023brief,li2024llava}, the researcher~\cite{lee2023multimodal} attempts to leverage the strong robustness of LLMs benefit from extensive training data to alleviate performance decreases caused by incomplete modalities. They design some sophisticated prompts to inform LLMs of different distributions of inputs. However, they only capture modality-common information as well (Fig.~\ref{fig:intro}a), ignoring distinct modality-specific knowledge of the missing modality. As such, existing methods still struggle to effectively incorporate the learning of both common and specific knowledge of missing modalities.

Based on these insights, to bridge the aforementioned gap, we propose a \textbf{Dis}tilled \textbf{Pro}mpt Learning (DisPro) framework to unlock LLMs' robustness to incomplete modalities, which aims to simultaneously compensate for both specific and common information of missing modalities. It comprises two-stage prompt learning including Unimodal Prompting (UniPro) and Multimodal Prompting (MultiPro) for capturing modality-specific and modality-common information of missing modalities, respectively.

In the first stage, to prepare for supplementing modality-specific information of missing modalities, we first learn the knowledge distribution of the missing modality by distilling unimodal knowledge into a set of learnable unimodal prompts for each modality from all remaining samples available of this modality. Specifically, inspired by CoOp~\cite{zhou2022learning}, a set of learnable prompts for different classes is employed to learn knowledge about how to describe different categories of an individual modality. For example, in survival prediction, learned prompts can be intuitively interpreted as how different risk groups will reflect in pathological morphology or gene expression profiles. However, due to the huge size (e.g., 100,000 $\times$ 100,000) of pathological WSIs, we cannot simply apply CoOp designed for small natural images to WSIs. Therefore, in UniPro, we extend CoOp into the multiple instance learning (MIL) paradigm that is commonly used for large-scale WSIs~\cite{ilse2018attention,yang2024mambamil,gadermayr2024multiple}.

In the second stage, to capture modality-common information for missing modalities, we leverage available modalities as multimodal prompts of LLMs to infer representations of missing modalities using LLMs' strong inference ability. Simultaneously, to inject the learned unimodal knowledge in stage 1, we propose UniPro Distillation to enforce the inferred representations of missing modalities to get as close as possible to the learned unimodal prompts of the corresponding class. Additionally, to assist in learning modality-common knowledge from WSIs and genes with high information redundancy~\cite{zhangprototypical,song2024multimodal}, we propose UniPro Scoring to re-use UniPro as a grader to select discriminative and relevant tokens. As a result, both common and specific information are compensated for missing modalities under the collaboration of uni- and multimodal prompts, thus mitigating performance declines led by incomplete modalities.

It is worth noting that DisPro is proposed as a framework for addressing missing modality issues by leveraging LLMs' power based on prompts, capable of generalizing to a wider range of domains. The contributions of this work are summarized as follows: 
\begin{itemize}
    \item We propose a Distilled Prompt Learning (DisPro) framework for incomplete multimodal survival prediction, where LLMs' robustness to incomplete modalities is explored under two-stage prompting.
    \item We propose two-stage prompt learning including Unimodal Prompting and Multimodal Prompting to compensate for both modality-specific and modality-common information of missing modalities, respectively.
    \item Extensive experiments on five benchmark survival prediction datasets are conducted, which demonstrates significantly superior to state-of-the-art approaches.
\end{itemize}
\section{Related Work}
\label{sec:related}

\subsection{Incomplete Multimodal Learning}
As stated above, incomplete multimodal learning~\cite{zhou2024multimodal} can be categorized into two types: imputation-based~\cite{hamghalam2021modality,yuan2023rethinking,chen2020hgmf} and imputation-free methods~\cite{ning2021relation, liu2022moddrop++}. For example, with the success of the diffusion model~\cite{croitoru2023diffusion}, the research~\cite{meng2024multi} attempted to apply it to MRI synthesis for missing modalities. Instead of generating raw modality data, SMIL~\cite{ma2021smil} leveraged Bayesian meta-learning to impute features of missing modalities. For retrieval-based approaches, M3Care~\cite{zhang2022m3care} filled in the missing modalities by retrieving the most similar samples from the training set. On the other hand, imputation-free approaches provided more flexible solutions, which aim to enhance the robustness of fusion and alleviate performance decrease. ~\cite{ding2021rfnet} proposed a segmentation-based regularizer to reduce the sensitivity to missing modalities. MUSE~\cite{wu2024multimodal} utilized patients' relationships based on graph networks to perform contrastive learning for improving the robustness to missing modalities. GTP-4o~\cite{li2024gtp} proposed modality-prompted heterogeneous graph to embed each modality into a unified space and complete missing modality through a graph prompting mechanism. Recently, MAP~\cite{lee2023multimodal} proposed missing-aware prompts to unlock the robustness of LLMs to missing modalities by indicating LLMs different missing scenarios. However, these methods typically supplemented modality-common knowledge for missing modalities.

\subsection{Multimodal Survival Prediction}
With the success of digital pathology~\cite{niazi2019digital,li2021dual,zhang2022dtfd} and sequencing techniques~\cite{ozsolak2011rna,stark2019rna}, pathology and genomic data are commonly integrated to analyze survival outcomes, which have proven the superiority on precision in previous works~\cite{chen2021multimodal,xu2023multimodal}. MCAT~\cite{chen2021multimodal} proposed to leverage the interactions across modalities to guide the learning of pathology data. Based on this work, MOTCat~\cite{xu2023multimodal} further leveraged optimal transport to capture global structure consistency across modalities. In parallel, CMTA~\cite{zhou2023cross} investigated the intrinsic cross-modal interactions and potential complementary information by cross-modal translation and alignment. From the perspective of information bottleneck, PIBD~\cite{zhangprototypical} proposed a prototypical information bottleneck for the MIL setting to select discriminative instances and disentangle multimodal knowledge for redundancy reduction. Similarly, MMP~\cite{song2024multimodal} utilized Gaussian Mixture Model to summarize morphological prototypes and biological pathway prototypes for these two modalities. However, all these methods assume all modalities available, inevitably leading to the difficulty of employment in clinical settings.

\begin{figure*}[t]
    \centering
    \includegraphics[width=0.95\linewidth]{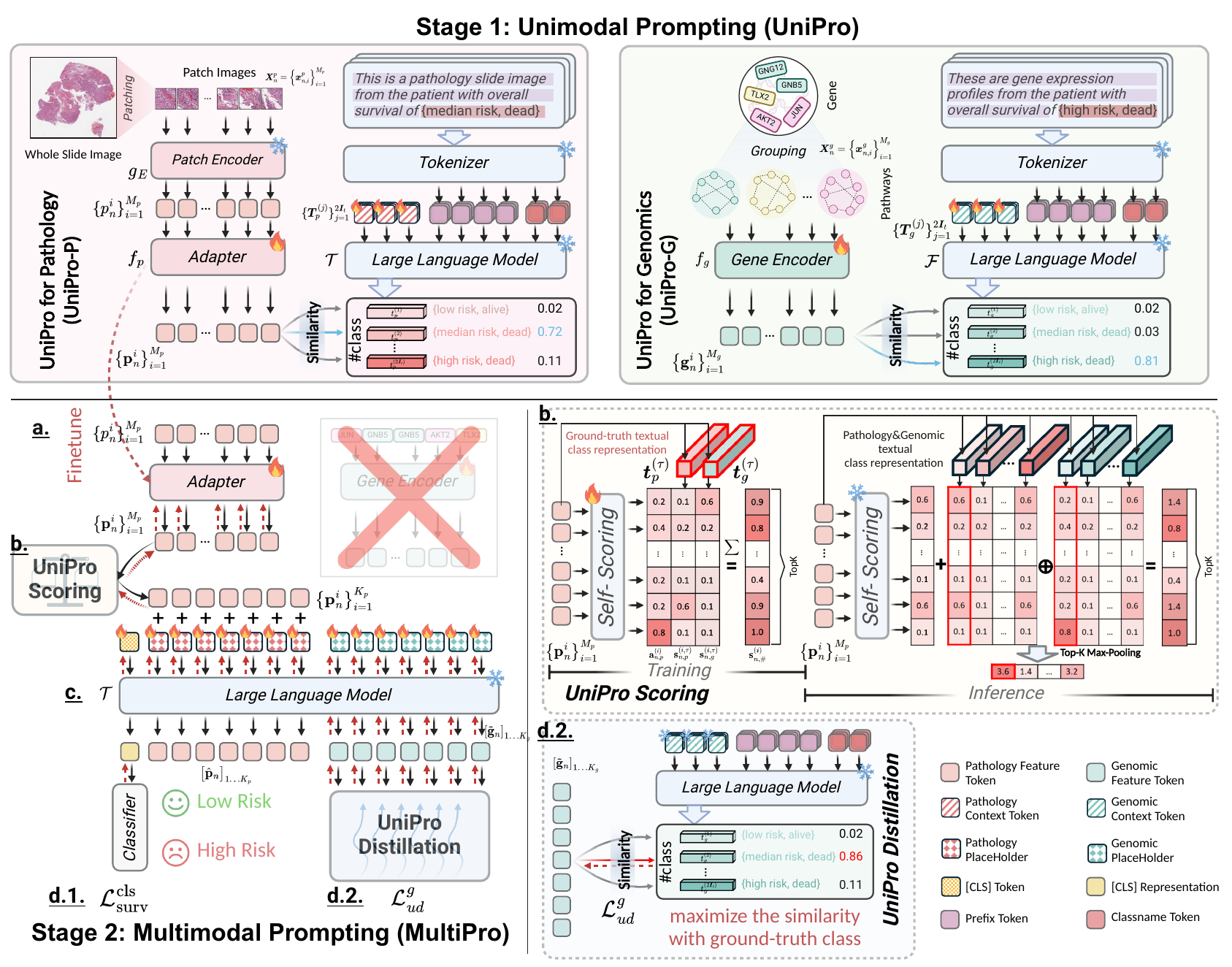}
    \vspace{-0.3cm}
    \caption{\textbf{Overview of DisPro}. Stage 1: Unimodal Prompting aims to distill the knowledge distribution for each modality and prepare to supplement modality-specific knowledge for the missing modality. Stage 2: Multimodal Prompting utilizes the available modality as prompts to infer the representations of the missing one, compensating for modality-common knowledge. Simultaneously, the learned UniPro of Stage 1 supervises the learning of imputed representations to compensate for modality-specific knowledge. UniPro Scoring re-uses learned prompts to assist the LLM in capturing modality-common knowledge by selecting discriminative and relevant tokens.}
    \vspace{-0.3cm}
    \label{fig:overview}
\end{figure*}

\section{Method}
\label{sec:method}
\label{sec:method_overview}
This work aims to explore LLMs' robustness to incomplete modalities by two-stage prompting, Unimodal Prompting (UniPro) and Multimodal Prompting (MultiPro), which capture modality-specific and modality-common information, respectively. The overview framework of the proposed method is shown in Fig.~\ref{fig:overview}. First, we will introduce the problem formulation in Sec.~\ref{sec:method_formulate}. Then, we will detail the process of UniPro and MultiPro in Sec.~\ref{sec:method_uni} and ~\ref{sec:method_multi}, respectively.
\subsection{Problem Formulation}
\label{sec:method_formulate}
Given a multimodal dataset consisting of $M=2$ modalities, we use a 2-tuple $\boldsymbol{X}_n=(\boldsymbol{X}^p_n, \boldsymbol{X}^g_n)$ to represent the $i$-th patient data, where $\boldsymbol{X}^p_n$ and $\boldsymbol{X}^g_n$ are pathology and genomic data, respectively. In particular, we use the notation of tilde hat to indicate the modality is missing, e.g., $\widetilde{\boldsymbol{X}}^p_n$.

Following the conventional setting for pathological WSIs and genomic pathways in previous works~\cite{jaume2024modeling,zhangprototypical}, we formulate a WSI $\boldsymbol{X}^p_n$ and a set of genomic pathways $\boldsymbol{X}^g_n$ as the ``bag'' of instances in MIL paradigm. We denote $\boldsymbol{X}^p_n=\{ \boldsymbol{x}^p_{n,i}\}_{i=1}^{M_p}$ where each instance $\boldsymbol{x}^p_{n,i}$ refers to a patch image cropped from the WSI $\boldsymbol{X}^p_n$ and $M_p$ is the patch number of this WSI. Similarly, the bag of genomic pathways can be formulated as $\boldsymbol{X}^g_n=\{ \boldsymbol{x}^g_{n,i}\}_{i=1}^{M_g}$, in which an instance $\boldsymbol{x}^g_{n,i}$ represents a pathway consisting of various gene expression values.

In survival prediction, the goal is to estimate the risk probability of an outcome event taking place before a specific time. However, this event may not always be observed, resulting in a right-censored occurrence. Therefore, $c_n\in \{0, 1\}$ is used to denote censorship status indicating whether the outcome event is right-censored (i.e., alive) $(c = 1)$ or not $(c = 0)$ (i.e., dead), and $t_n \in \mathbb{R}^+$ represents the overall survival time (in months). These two components make up its set of labels $\boldsymbol{Y}_n=(c_n, t_n)$. Following the previous setting~\cite{chen2021multimodal,xu2023multimodal}, the continuous survival time would be discretized into $\boldsymbol{I}_t$ time intervals corresponding to $\boldsymbol{I}_t$ risk bands. As a result, the label set $(c_n, t_n)$ can be re-written as $y_n$, resulting in $2\boldsymbol{I}_t$ labels.

Given a set of predicted hazard probabilities $\textbf{h}_n$ across all time intervals, the hazard probability $\textbf{h}^{(j)}_n$ represents the probability of death event occurring at the $j$-th time interval. Then, the cumulative survival probability for each time interval can be estimated as follows:
\begin{equation}
    S_n^{(j)} = \prod_{z=1}^{j}(1-\textbf{h}_n^{(z)})
    \vspace{-0.2cm}
\end{equation}
Following previous works~\cite{zhangprototypical,xu2023multimodal}, given the ground truth time interval $\tau$, we use the NLL loss~\cite{zadeh2020bias} for survival prediction, where $N$ is the sample number of the training set:
\begin{equation}
\label{eq:surv}
\begin{aligned}
    \mathcal{L}_{surv} = -\sum_{n=1}^N [ c_n \log S_n^{(\tau)} &+ (1-c_n) \log S_n^{(\tau-1)} \\
    &+ (1-c_n) \log \textbf{h}_n^{(\tau)}]
\end{aligned}
\end{equation}
\subsection{Stage 1: Unimodal Prompting (UniPro)}
\label{sec:method_uni}
Given a WSI $\boldsymbol{X}^p_n=\{ \boldsymbol{x}^p_{n,i}\}_{i=1}^{M_p}$ or a bag of genomic pathways $\boldsymbol{X}^g_n=\{ \boldsymbol{x}^g_{n,i}\}_{i=1}^{M_g}$, we formulate their learning process as a MIL task. Inspired by CoOp~\cite{zhou2022learning} employing a CLIP-style architecture, we adapt it to the MIL setting to learn how to characterize each modality using a set of learnable prompts for modality-specific knowledge.

In terms of pathological WSIs data, following the common setting of the WSI processing~\cite{shao2021transmil}, as shown in the UniPro-P of Fig~\ref{fig:overview}, we first employ a pretrained and frozen patch encoder $g_E(\cdot)$ to embed every patch images $\boldsymbol{x}^p_{n,i}$ from a WSI into an instance-level embedding $p_n^i=g_E(\boldsymbol{x}^p_{n,i}) \in \mathbb{R}^d$, where $d$ is the patch feature dimension. To align pathology features with the text space, we apply an adapter $f_p(\cdot)$ for each patch feature to get the aligned patch feature $\textbf{p}_n^i=f_p(g_E(\boldsymbol{x}^p_{n,i})) \in \mathbb{R}^D$, implemented by a linear layer followed by ReLU~\cite{brownlee2019gentle}.

Text-wise, the input comprises three components: pathology context tokens, prefix tokens and classname tokens. For classname tokens $[C]_{1\ldots s}$, we map every time interval of overall survival into a unique description, such as ``\textit{median risk}'', and denote the censorship status by ``\textit{alive}'' or ``\textit{dead}''. The prefix tokens $\text{[PREFIX]}_{1\ldots m}$ briefly describe the modality, e.g., ``\textit{This is a pathology slide image from the patient with overall survival of }$[C]_{1\ldots s}$''. Prefix and classname tokens are obtained using a frozen Tokenizer pretrained on a large-scale biomedical corpus to embed these words. To learn how to describe the modality for a specific risk band (class), we attach a set of learnable context tokens $[P]_{1\ldots k} $ to prefix and classname tokens:
\begin{equation}
\label{eq:tp}
\small
    \boldsymbol{T}_p=[P]_1\ldots [P]_k\text{[PREFIX]}_1\ldots\text{[PREFIX]}_m [C]_1\ldots [C]_s
    \vspace{-0.1cm}
\end{equation}
where $k$, $m$ and $s$ are the token numbers of pathology context tokens, prefix words and classname words, respectively. Each token is a vector with the same dimension $D$ as word embeddings of the LLM (i.e., 768 for BERT). In this way, we can design input tokens for each risk band:
\begin{equation}
\label{eq:tp_j}
\small
    \boldsymbol{T}_p^{(j)}=[P]_1^{(j)}\ldots [P]_k^{(j)}\text{[PREFIX]}_1\ldots\text{[PREFIX]}_m [C]_1^{(j)}\ldots [C]_s^{(j)}
\end{equation}
in which $j \in [1, 2\boldsymbol{I}_t]$ and prefix tokens are shared across all risk bands. By forwarding $\boldsymbol{T}_p^{(j)}$ of each class to the text encoder $\mathcal{T}(\cdot)$ (i.e., LLM), we can obtain a global textual class representation $\boldsymbol{t}_p^{(j)} \in \mathbb{R}^D$for every class.

In the MIL setting, to obtain the predicted probability for each class, we can use topK patches to produce the slide-level prediction based on similarity scores. To this end, we first compute the similarity score between every instance and every class:
\begin{equation}
\label{eq:uni_sim}
    \boldsymbol{s}_{n,p}^{(i, j)}=\textbf{p}_n^{i}\cdot \boldsymbol{t}_p^{(j)}
    \vspace{-0.1em}
\end{equation}
As a result, we can get a similarity matrix $\mathcal{S}_{n,p} \in \mathbb{R}^{M_p \times 2\boldsymbol{I}_t}$for a WSI. Then, we apply a TopK max-pooling operator $h(\cdot)$ to obtain the slide-level predictions, depending on top-K similarities for each class:
\begin{equation}
\label{eq:topk_max}
    h(\mathcal{S}_{n,p}) = \frac{1}{K}\left[ \sum_{i=1}^K \hat{\boldsymbol{s}}_{n,p}^{(i, 1)}, \sum_{i=1}^K \hat{\boldsymbol{s}}_{n,p}^{(i, 2)},\ldots, \sum_{i=1}^K \hat{\boldsymbol{s}}_{n,p}^{(i, 2\boldsymbol{I}_t)} \right]
\end{equation}
where $\hat{\boldsymbol{s}}_{n,p}$ is the sorted $\boldsymbol{s}_{n,p}$ in a descending order across all instances. Next, we can get a set of hazard probabilities $\textbf{h}_n^p$ across all time intervals through a sigmoid function.
\begin{equation}
\label{eq:stage1_hazard}
    \textbf{h}_n^p = \text{Sigmoid}(h(\mathcal{S}_{n,p}))
\end{equation}
By computing the survival loss according to Eq.~\ref{eq:surv} and back-forwarding gradients, we can optimize the modality adapter and learnable prompts consisting of context tokens.

Similarly, we acquire the set of hazard probabilities $\textbf{h}_n^g$ for genomic pathways and optimize its learnable prompts $\boldsymbol{T}_g$ and gene encoder $f_g(\cdot)$. Here we use the subscript $p$ and $g$ to represent pathology and genomic data. Every pathway is also encoded into a $D$-dimensional embedding.
\subsection{Stage 2: Multimodal Prompting (MultiPro)}
\label{sec:method_multi}
In this stage, we aim to leverage LLMs' strong inference ability and robustness to missing modalities to predict the representation of missing modalities, as shown in the MultiPro of Fig.~\ref{fig:overview}.

To simplify explanations, we take the example input of pathology available and genomics being missing. By prompting the LLM with available pathology data, the Bert-style LLM can capture modality-common knowledge for genomics due to its nature of self-attention. To compensate for modality-specific knowledge, we re-use the unimodal prompts $\boldsymbol{T}_g$ learned in stage 1 with all remaining samples available of genomics data to supervise the learning of imputed representations, where unimodal prompts characterize the knowledge distribution of an individual modality. 

To this end, we first need to construct an interleaved input for the learning of modality-common information. To fully leverage the capabilities of LLMs, every modality are supposed to be embedded into the aligned text space. Consequently,  we re-use the aligned modality adapter or encoder learned in the first stage to embed each modality, and we continue to finetune them in the second stage. Note that the LLM used in stage 2 is the same as the one in stage 1. As a result, we can get a bag of aligned feature tokens $\{ \textbf{p}_n^i \}_{i=1}^{M_p}$ or $\{ \textbf{g}_n^i \}_{i=1}^{M_g}$ for pathology and genomic
data, respectively. Here we assume genomic data $\{ \tilde{\textbf{g}}_n^i \}_{i=1}^{M_g}$ is missing.

Due to the limitation of input length for general LLMs (e.g., 512 for Bert-style LLMs) and high information redundancy in WSIs or genomic pathways~\cite{zhangprototypical,song2024multimodal}, more discriminative and relevant feature tokens are expected to be selected for better capturing modality-common knowledge. Therefore, we propose a UniPro Scoring module by re-using UniPro as a grader to estimate a relevance score for each feature token.

\noindent
\textbf{UniPro Scoring.} The score of every feature token is determined by three parts: relevance scores estimated by UniPro-P and UniPro-G, and self-scoring, as shown in Fig.~\ref{fig:overview}.

Given a set of pathology feature tokens $\{ \textbf{p}_n^i \}_{i=1}^{M_p}$, to obtain discriminative tokens, we compute the similarity score $\boldsymbol{s}_{n,p}^{(i, \tau)}$ between the ground-truth textual class representation of $\boldsymbol{t}_p^{(\tau)}$ and each feature token $\textbf{p}_n^i$ by re-using Eq.~\ref{eq:uni_sim}. Through the sigmoid function, we can get the score $\textbf{s}_{n, p}^{(i, \tau)}$ estimated by UniPro-P. To capture cross-modal modality-common information for better compensation of missing modality, we similarly obtain the score $\textbf{s}_{n,g}^{(i, \tau)}$ estimated by UniPro-G. Since textual class representations of UniPro are fixed once finishing the training of stage 1, the selected tokens will also become fixed accordingly. To make it diverse while dynamically adapting to the current input, we employ a learnable self-scoring mechanism implemented by attention layers~\cite{ilse2018attention} to assign a score $\textbf{a}_{n,p}^{(i)}$ for each feature token. In the end, the final score $\textbf{s}_{n,\#}^{(i)}$ for each feature token can be computed by their summarization:
\begin{equation}
\label{eq: stage2_scores_sum}
    \textbf{s}_{n,\#}^{(i)} = \textbf{s}_{n,p}^{(i, \tau)} + \textbf{s}_{n,g}^{(i, \tau)} + \textbf{a}_{n,p}^{(i)}
    \vspace{-0.2cm}
\end{equation}
For the inference phase when the ground-truth class cannot be accessed, we compute similarity scores with all textual class representations and apply Top-K max-pooling to determine which class $j$ these tokens belong to:
\begin{equation}
    \tau = \arg\max_{j\in [1, 2\boldsymbol{I}_t]} h(\mathcal{S}_{n,p}^{(j)}\oplus\mathcal{S}_{n,g}^{(j)})
    \vspace{-0.2cm}
\end{equation}
where $\oplus$ refers to the element-wise addition and $h(\cdot)$ can be computed by Eq.~\ref{eq:topk_max}.

By selecting feature tokens with the top-$K_p$ largest scores, we can get the input tokens $\{\textbf{p}_n^i \}_{i=1}^{K_p}$ for available pathology modality. Then, we employ a set of learnable tokens to act as placeholders $[H_p]_{1\ldots K_p}$ and $[H_g]_{1\ldots K_g}$ for pathology and genomic data, where $K_p$ and $K_g$ are their token number of the input for the LLM in stage 2, respectively. They are initialized with modality indicators. For the available pathology, we add the selected tokens to its placeholders, leading to the input of pathology $\{\hat{\textbf{p}}_n^i \}_{i=1}^{K_p}$. At last, we concatenate them with placeholders of the missing genomics as well as a [CLS] token, resulting in the final input being forwarded into a LLM $\mathcal{T}(\cdot)$:
\begin{equation}
\label{eq:stage2_out}
\small
    [CLS][\hat{\textbf{p}}_n]_{1\ldots K_p}[\tilde{\textbf{g}}_n]_{1\ldots K_g}=\mathcal{T}([\text{CLS}]||\{\hat{\textbf{p}}_n^i \}_{i=1}^{K_p}||[H_g]_{1\ldots K_g})
\end{equation}

\noindent
\textbf{UniPro Distillation.} To inject modality-specific knowledge for the compensation of missing modality, we utilize UniPro to supervise this process, where UniPro has learned the knowledge distribution regarding the missing modality based on other observed samples in stage 1.

Given the output (Eq.~\ref{eq:stage2_out}) of the LLM in stage 2, we take the part of the missing modality $[\tilde{\textbf{g}}_n]_{1\ldots K_g}$ and compute similarity scores $\tilde{\textbf{h}}_g$ between $[\tilde{\textbf{g}}_n]_{1\ldots K_g}$ and the textual class representation by Eq.~\ref{eq:uni_sim} - ~\ref{eq:stage1_hazard}. By substituting $[\tilde{\textbf{g}}_n]_{1\ldots K_g}$ into the survival loss function of Eq.~\ref{eq:surv}, we can supervise the learning of compensation for the missing modality:
\begin{equation}
    \label{eq:surv_genomic}
    \begin{aligned}
    \mathcal{L}_{ud}^g = -\sum_{n=1}^{\tilde{N}_g} [ c_n \log S_n^{(\tau)} &+ (1-c_n) \log S_n^{(\tau-1)} \\
    &+ (1-c_n) \log \tilde{\textbf{h}}_n^{g, (\tau)}]
    \end{aligned}
    \vspace{-0.3cm}
\end{equation}
where $\tilde{N}_g$ is the number of missing genomic modalities across all data pairs. Similarly, we can get the loss function $\mathcal{L}_{ud}^p$ when the pathology modality is missing. Lastly, $[CLS]$ token is used for calculating the loss function for survival prediction $\mathcal{L}_{surv}^{cls}$ in stage 2, and the final loss can be combined linearly with coefficient factors $\alpha_1$ and $\alpha_2$:
\begin{equation}
\mathcal{L} = \mathcal{L}_{surv}^{cls} + \alpha_1\mathcal{L}_{ud}^p + \alpha_2\mathcal{L}_{ud}^g
\end{equation}
\section{Experiments}
\label{sec:exp}
\begin{table*}[htbp]
  \centering
  \scalebox{0.85}{
    \begin{tabular}{c|cc|l|ccccc|c}
    \toprule
    \multicolumn{1}{c|}{\multirow{2}[4]{*}{\makecell[c]{Training\\MiRate}}} & \multicolumn{2}{c|}{Test} & \multirow{2}[4]{*}{Methods} & \multicolumn{1}{c}{\multirow{2}[4]{*}{\makecell[c]{BLCA\\(372)}}} & \multicolumn{1}{c}{\multirow{2}[4]{*}{\makecell[c]{BRCA\\(1007)}}} & \multicolumn{1}{c}{\multirow{2}[4]{*}{\makecell[c]{COADREAD\\(533)}}} & \multicolumn{1}{c}{\multirow{2}[4]{*}{\makecell[c]{LUAD\\(443)}}} & \multicolumn{1}{c|}{\multirow{2}[4]{*}{\makecell[c]{UCEC\\(478)}}} & \multirow{2}[4]{*}{Avg} \\
\cmidrule{2-3}          & P     & G     &       &       &       &       &       &       &  \\
    \midrule
    \multicolumn{3}{c|}{\multirow{6}[6]{*}{N/A}} & $\dagger$TransMIL~\cite{shao2021transmil} & 0.5903±0.025 & 0.6671±0.054 & 0.6093±0.018 & 0.6195±0.035 & \textbf{0.6978±0.046} & 0.6368 \\
    \multicolumn{3}{c|}{} & $\dagger$CoOp\_BioBert~\cite{zhou2022learning} & \textbf{0.6014±0.047} & \textbf{0.6784±0.032} & \textbf{0.6681±0.026} & \textbf{0.6543±0.052} & 0.6791±0.085 & \textbf{0.6563} \\
\cmidrule{4-10}    \multicolumn{3}{c|}{} & $\dagger$SNN~\cite{klambauer2017self}   & \textbf{0.6831±0.040} & 0.6567±0.038 & 0.6312±0.058 & 0.6379±0.071 & \textbf{0.7267±0.073} & 0.6671 \\
    \multicolumn{3}{c|}{} & $\dagger$CoOp\_BioBert~\cite{zhou2022learning} & 0.6712±0.036 & \textbf{0.6922±0.035} & \textbf{0.6743±0.048} & \textbf{0.6514±0.065} & 0.7171±0.060 & \textbf{0.6812} \\
\cmidrule{4-10}    \multicolumn{3}{c|}{} & $\ddagger$MOTCat~\cite{xu2023multimodal} & 0.6274±0.046 & 0.6721±0.038 & 0.6501±0.035 & 0.6753±0.043 & 0.7207±0.074 & 0.6691 \\
    \multicolumn{3}{c|}{} & $\ddagger$SurvPath~\cite{jaume2024modeling} & \textbf{0.6572±0.027} & \textbf{0.7071±0.039} & \textbf{0.7079±0.048} & \textbf{0.6798±0.049} & \textbf{0.7391±0.065} & \textbf{0.6982} \\
    \midrule
    \midrule
    \multirow{18}[6]{*}{60\%} & $\bullet$ & $\circ$ & COM~\cite{qian2023contrastive}   & 0.6024±0.013 & 0.6740±0.011 & 0.6779±0.014 & 0.6336±0.011 & 0.6985±0.012 & 0.6573 \\
          & $\bullet$ & $\circ$ & M3Care~\cite{zhang2022m3care} & \underline{0.6208±0.014} & 0.6690±0.018 & 0.6570±0.023 & 0.6219±0.007 & 0.7031±0.012 & 0.6543 \\
          & $\bullet$ & $\circ$ & HGCN~\cite{hou2023hybrid}  & 0.5999±0.009 & \underline{0.6831±0.012} & \underline{0.6792±0.009} & 0.6465±0.005 & 0.7011±0.007 & \underline{0.6620} \\
          & $\bullet$ & $\circ$ & MUSE~\cite{wu2024multimodal}  & 0.6207±0.013 & 0.6514±0.014 & 0.6547±0.009 & 0.6354±0.012 & \underline{0.7048±0.023} & 0.6534 \\
          & $\bullet$ & $\circ$ & MAP~\cite{lee2023multimodal}   & 0.5918±0.012 & 0.6278±0.006 & 0.5965±0.017 & \underline{0.6489±0.020} & 0.6929±0.021 & 0.6316 \\
          \rowcolor{red!10}
          \cellcolor{white}& $\bullet$ & $\circ$ & DisPro (Ours) & \textbf{0.6319±0.014} & \textbf{0.6895±0.011} & \textbf{0.6880±0.010} & \textbf{0.6612±0.014} & \textbf{0.7272±0.018} & \textbf{0.6796} \\
\cmidrule{2-10}          & $\circ$ & $\bullet$ & COM~\cite{qian2023contrastive}   & 0.6415±0.016 & 0.6653±0.018 & 0.6593±0.030 & 0.6456±0.008 & \underline{0.7105±0.007} & \underline{0.6645} \\
          & $\circ$ & $\bullet$ & M3Care~\cite{zhang2022m3care} & \underline{0.6420±0.015} & 0.6642±0.029 & 0.6491±0.008 & 0.6462±0.012 & 0.7032±0.013 & 0.6609 \\
          & $\circ$ & $\bullet$ & HGCN~\cite{hou2023hybrid}  & 0.6225±0.008 & \underline{0.6729±0.009} & \underline{0.6752±0.015} & 0.6418±0.015 & 0.7015±0.010 & 0.6628 \\
          & $\circ$ & $\bullet$ & MUSE~\cite{wu2024multimodal}  & 0.6299±0.012 & 0.6427±0.025 & 0.6391±0.023 & \underline{0.6462±0.010} & 0.6893±0.027 & 0.6494 \\
          & $\circ$ & $\bullet$ & MAP~\cite{lee2023multimodal}   & 0.5937±0.016 & 0.6269±0.006 & 0.5971±0.017 & 0.6442±0.009 & 0.6934±0.018 & 0.6311 \\
          \rowcolor{red!10}
          \cellcolor{white}& $\circ$ & $\bullet$ & DisPro (Ours) & \textbf{0.6547±0.012} & \textbf{0.6841±0.018} & \textbf{0.6804±0.024} & \textbf{0.6548±0.012} & \textbf{0.7271±0.017} & \textbf{0.6802} \\
\cmidrule{2-10}          & $\bullet$ & $\bullet$ & COM~\cite{qian2023contrastive}   & 0.6192±0.013 & 0.6846±0.009 & \underline{0.6874±0.005} & 0.6548±0.012 & 0.7136±0.016 & 0.6719 \\
          & $\bullet$ & $\bullet$ & M3Care~\cite{zhang2022m3care} & \underline{0.6513±0.005} & \underline{0.6951±0.024} & 0.6698±0.016 & 0.6491±0.013 & \underline{0.7288±0.011} & \underline{0.6788} \\
          & $\bullet$ & $\bullet$ & HGCN~\cite{hou2023hybrid}  & 0.6332±0.006 & 0.6907±0.012 & 0.6861±0.017 & 0.6545±0.011 & 0.7077±0.008 & 0.6745 \\
          & $\bullet$ & $\bullet$ & MUSE~\cite{wu2024multimodal}  & 0.6207±0.012 & 0.6555±0.013 & 0.6539±0.012 & 0.6415±0.008 & 0.7164±0.013 & 0.6576 \\
          & $\bullet$ & $\bullet$ & MAP~\cite{lee2023multimodal}   & 0.5906±0.013 & 0.6277±0.004 & 0.5990±0.016 & \underline{0.6625±0.008} & 0.6901±0.018 & 0.6340 \\
          \rowcolor{red!10}
          \cellcolor{white}& $\bullet$ & $\bullet$ & DisPro (Ours) & \textbf{0.6638±0.006} & \textbf{0.7219±0.015} & \textbf{0.7029±0.016} & \textbf{0.6741±0.007} & \textbf{0.7476±0.020} & \textbf{0.7021} \\
    \midrule
    \multirow{18}[6]{*}{0\%} & $\bullet$ & $\circ$ & COM~\cite{qian2023contrastive}   & 0.6154±0.041 & \underline{0.6871±0.027} & \underline{0.6925±0.044} & 0.6427±0.046 & 0.6806±0.061 & \underline{0.6637} \\
          & $\bullet$ & $\circ$ & M3Care~\cite{zhang2022m3care} & \underline{0.6268±0.024} & 0.6775±0.016 & 0.6850±0.052 & 0.6299±0.038 & 0.6964±0.061 & 0.6631 \\
          & $\bullet$ & $\circ$ & HGCN~\cite{hou2023hybrid}  & 0.6055±0.048 & 0.6716±0.022 & 0.6734±0.048 & 0.6421±0.052 & 0.6915±0.053 & 0.6568 \\
          & $\bullet$ & $\circ$ & MUSE~\cite{wu2024multimodal}  & 0.6208±0.026 & 0.6767±0.040 & 0.6710±0.041 & 0.6253±0.041 & \underline{0.7099±0.063} & 0.6607 \\
          & $\bullet$ & $\circ$ & MAP~\cite{lee2023multimodal}   & 0.6205±0.033 & 0.6300±0.053 & 0.5930±0.065 & \underline{0.6662±0.055} & 0.6925±0.064 & 0.6404 \\
          \rowcolor{red!10}
          \cellcolor{white}& $\bullet$ & $\circ$ & DisPro (Ours) & \textbf{0.6315±0.028} & \textbf{0.6945±0.044} & \textbf{0.7216±0.028} & \textbf{0.6747±0.052} & \textbf{0.7104±0.056} & \textbf{0.6865} \\
\cmidrule{2-10}          & $\circ$ & $\bullet$ & COM~\cite{qian2023contrastive}   & 0.6681±0.030 & \underline{0.6966±0.047} & 0.6770±0.024 & 0.6454±0.050 & \underline{0.7095±0.059} & \underline{0.6793} \\
          & $\circ$ & $\bullet$ & M3Care~\cite{zhang2022m3care} & \underline{0.6706±0.039} & 0.6611±0.040 & 0.6563±0.021 & 0.6512±0.061 & 0.7047±0.060 & 0.6688 \\
          & $\circ$ & $\bullet$ & HGCN~\cite{hou2023hybrid}  & 0.6233±0.033 & 0.6807±0.023 & \underline{0.6777±0.037} & 0.6421±0.045 & 0.7091±0.043 & 0.6666 \\
          & $\circ$ & $\bullet$ & MUSE~\cite{wu2024multimodal}  & 0.6256±0.029 & 0.6401±0.024 & 0.6208±0.030 & 0.6547±0.075 & 0.6824±0.039 & 0.6447 \\
          & $\circ$ & $\bullet$ & MAP~\cite{lee2023multimodal}   & 0.6093±0.026 & 0.6305±0.051 & 0.5842±0.065 & \underline{0.6653±0.045} & 0.6841±0.064 & 0.6347 \\
          \rowcolor{red!10}
          \cellcolor{white}& $\circ$ & $\bullet$ & DisPro (Ours) & \textbf{0.6745±0.039} & \textbf{0.6996±0.028} & \textbf{0.7000±0.021} & \textbf{0.6751±0.068} & \textbf{0.7140±0.060} & \textbf{0.6926} \\
\cmidrule{2-10}          & $\bullet$ & $\bullet$ & COM~\cite{qian2023contrastive}   & 0.6376±0.043 & \underline{0.7087±0.024} & 0.6990±0.045 & 0.6659±0.034 & 0.7112±0.051 & 0.6845 \\
          & $\bullet$ & $\bullet$ & M3Care~\cite{zhang2022m3care} & \underline{0.6605±0.023} & 0.6808±0.041 & 0.6768±0.026 & 0.6740±0.046 & \textbf{0.7359±0.033} & \underline{0.6856} \\
          & $\bullet$ & $\bullet$ & HGCN~\cite{hou2023hybrid}  & 0.6365±0.021 & 0.7058±0.033 & \underline{0.7136±0.025} & 0.6692±0.050 & 0.7017±0.044 & 0.6854 \\
          & $\bullet$ & $\bullet$ & MUSE~\cite{wu2024multimodal}  & 0.6282±0.039 & 0.7017±0.020 & 0.6693±0.042 & 0.6248±0.040 & \underline{0.7237±0.062} & 0.6695 \\
          & $\bullet$ & $\bullet$ & MAP~\cite{lee2023multimodal}   & 0.6034±0.026 & 0.6328±0.049 & 0.5895±0.063 & \underline{0.6904±0.047} & 0.6827±0.065 & 0.6398 \\
          \rowcolor{red!10}
          \cellcolor{white}& $\bullet$ & $\bullet$ & DisPro (Ours) & \textbf{0.6719±0.051} & \textbf{0.7358±0.051} & \textbf{0.7553±0.021} & \textbf{0.6767±0.041} & 0.7169±0.071 & \textbf{0.7113} \\
    \bottomrule
    \end{tabular}%
    }
    \vspace{-0cm}
    \caption{C-Index of Comparison to SOTA models. “MiRate” represents Missing Rate. ``P'' and ``G'' indicate Pathology and Genomics, respectively. $\circ$ and  $\bullet$ mean the missing modality and the available modality, respectively. $\dagger$ and $\ddagger$ indicate using unimodal and complete multimodal data, respectively, while the rest are for missing modalities. The best results are in \textbf{bold}. The second-best results are \underline{underlined}.}
    \vspace{-0.3cm}
  \label{tab:comp}%
\end{table*}%

\begin{table*}[htbp]
  \centering
  \scalebox{0.9}{
    \begin{tabular}{c|cc|cc|ccccc|c}
    \toprule
    \multirow{2}[4]{*}{Variants} & \multicolumn{2}{c}{Test} & \multicolumn{2}{c|}{Modules} & \multicolumn{1}{c}{\multirow{2}[4]{*}{\makecell[c]{BLCA\\(372)}}} & \multicolumn{1}{c}{\multirow{2}[4]{*}{\makecell[c]{BRCA\\(1007)}}} & \multicolumn{1}{c}{\multirow{2}[4]{*}{\makecell[c]{COADREAD\\(533)}}} & \multicolumn{1}{c}{\multirow{2}[4]{*}{\makecell[c]{LUAD\\(443)}}} & \multicolumn{1}{c|}{\multirow{2}[4]{*}{\makecell[c]{UCEC\\(478)}}} & \multirow{2}[4]{*}{Avg} \\
\cmidrule{2-5}          & P     & G     & US    & UD    &       &       &       &       &       &  \\
    \midrule
    baseline & $\bullet$ & $\circ$ &  &  & 0.6068±0.021 & 0.6668±0.006 & 0.6583±0.012 & 0.6270±0.018 & 0.7081±0.019 & 0.6534 \\
    + UD & $\bullet$ & $\circ$ &  & \checkmark & 0.6213±0.010 & 0.6829±0.008 & 0.6740±0.022 & 0.6327±0.018 & 0.7152±0.011 & 0.6652 \\
    + US & $\bullet$ & $\circ$ & \checkmark &  & 0.6285±0.021 & 0.6884±0.016 & 0.6653±0.026 & 0.6546±0.013 & 0.7203±0.007 & 0.6714 \\
    \rowcolor{red!10}
    DisPro (full) & $\bullet$ & $\circ$ & \checkmark & \checkmark & \textbf{0.6319±0.014} & \textbf{0.6895±0.011} & \textbf{0.6880±0.010} & \textbf{0.6612±0.014} & \textbf{0.7272±0.018} & \textbf{0.6796} \\
    \midrule
    baseline & $\circ$ & $\bullet$ &  &  & 0.6435±0.016 & 0.6605±0.033 & 0.6601±0.018 & 0.6467±0.016 & 0.7127±0.007 & 0.6647 \\
    + UD & $\circ$ & $\bullet$ &  & \checkmark & 0.6473±0.017 & 0.6836±0.014 & 0.6707±0.022 & 0.6420±0.020 & 0.7226±0.013 & 0.6732 \\
    + US & $\circ$ & $\bullet$ & \checkmark &  & 0.6452±0.020 & 0.6666±0.017 & 0.6763±0.023 & 0.6471±0.023 & 0.7151±0.009 & 0.6701 \\
    \rowcolor{red!10}
    DisPro (full) & $\circ$ & $\bullet$ & \checkmark & \checkmark & \textbf{0.6547±0.012} & \textbf{0.6841±0.018} & \textbf{0.6804±0.024} & \textbf{0.6548±0.012} & \textbf{0.7271±0.017} & \textbf{0.6802} \\
    \midrule
    baseline & $\bullet$ & $\bullet$ &  &  & 0.6522±0.010 & 0.7075±0.012 & 0.6785±0.020 & 0.6541±0.009 & 0.7157±0.003 & 0.6816 \\
    + UD & $\bullet$ & $\bullet$ &  & \checkmark & 0.6585±0.007 & 0.7194±0.013 & 0.6963±0.009 & 0.6635±0.014 & 0.7250±0.012 & 0.6926 \\
    + US & $\bullet$ & $\bullet$ & \checkmark &  & 0.6593±0.008 & 0.7125±0.010 & 0.7005±0.018 & 0.6717±0.007 & 0.7274±0.010 & 0.6943 \\
    \rowcolor{red!10}
    DisPro (full) & $\bullet$ & $\bullet$ & \checkmark & \checkmark & \textbf{0.6638±0.006} & \textbf{0.7219±0.015} & \textbf{0.7029±0.016} & \textbf{0.6741±0.007} & \textbf{0.7476±0.020} & \textbf{0.7021} \\
    \bottomrule
    \end{tabular}%
    }
    \caption{\textbf{C-Index of Ablation Study on the Setting of 60\% Training Missing Rate}.``US'' and ``UD'' refer to the UniPro Scoring and UniPro Distillation modules, respectively. ``P'' and ``G'' indicate Pathology and Genomics, respectively. $\circ$ means the modality is missing, while $\bullet$ refers to the available modality.}
    \vspace{-0.3cm}
  \label{tab:ablation}%
\end{table*}%
\subsection{Datasets and Settings}
\textbf{Datasets}. To demonstrate the efficacy of the proposed method, we conducted a series of experiments across five public cancer datasets sourced from The Cancer Genome Atlas (TCGA)~\footnote{\url{https://portal.gdc.cancer.gov/}}. The case number of each dataset is demonstrated next to the dataset name in Tab.~\ref{tab:comp}. Regarding genomic data, we follow the grouping of biological pathways provided by ~\cite{jaume2024modeling} and remove genes absent in the cBioPortal~\footnote{\url{https://www.cbioportal.org/}} database, resulting in 330 pathways. More details of datasets can be found in supplementary materials.

\noindent
\textbf{Evaluation Settings on Incomplete Modalities.} Following the common setting in previous works~\cite{ma2021smil, wang2023multi}, to systematically assess performances on various missing settings, we first manually construct a series of missing scenarios under different missing ratios by randomly dropping them for training sets on samples with complete modality pairs. Specifically, we set a total missing rate (60\%) for two modalities and then construct 5 combinations for pathology and genomics, i.e., \{(0\%, 60\%), (20\%, 40\%), (30\%, 30\%), (40\%, 20\%), (60\%, 0\%)\}. For inference, all methods are evaluated on three scenarios: pathology only, genomics only, and complete modalities. All experiments are conducted on 5-fold cross-validation following previous works~\cite{chen2021multimodal,xu2023multimodal} in survival prediction, and the concordance index (C-Index) and its standard deviation (std) are reported to evaluate performances. The LLM is implemented by BioBERT-v1.2~\cite{lee2020biobert}. The supplementary materials contain more experiment results and details on the implementation.
\subsection{Comparisons with SOTAs}
We first present unimodal approaches, and SOTA multimodal methods trained with complete modalities (all pathology and genomics data) to serve as the upper bound of current complete modality approaches. Then the average performances on all missing combinations are demonstrated under the total training missing rate of 60\%, evaluated on three inference scenarios. Details on various combinations are shown in Fig.~\ref{fig:missing_curve} and we discuss them in Sec.~\ref{sec:ablation}. Lastly, we investigate the upper bound of each model tailored to incomplete modalities by training them with all data (without missing data).

From results in Tab.~\ref{tab:comp}, we observed that  (1) When applying 60\% training missing rate, DisPro consistently outperformed SOTAs across five datasets by a large margin (on average, +1.76\% for pathology only, +1.57\% for genomics only and 2.33\% for complete data). This indicates the superiority of DisPro over the incomplete SOTA methods. (2) When testing on the complete scenario, on 3 out of 5 datasets and overall, DisPro even surpassed the SOTA multimodal survival model (SurvPath) trained on all data. Notably, DisPro achieved this while only utilizing incomplete training data with a 60\% missing rate. It suggests that harnessing the powerful capabilities of LLMs is a promising solution for addressing the classical question of survival analysis. (3) When using complete training data, DisPro consistently performed best on average for various test scenarios as well (2.28\% on pathology only, 1.33\% on genomics only and 2.57\% on complete test data). Under the same condition, DisPro further increased by 1.31\% compared with the SOTA complete multimodal method, further demonstrating the strong power of LLMs. (4) In contrast to unimodal methods trained on full unimodal datasets, DisPro, using only incomplete multimodal data with a 60\% missing rate, achieved superior or similar performance when inferring from individual modalities. This showcases DisPro's efficient utilization of multimodal data.
\subsection{Ablation Study}
\label{sec:ablation}
\textbf{Component Validation}. In Tab.~\ref{tab:ablation}, we ablate each component in our method to investigate their effectiveness. (1) To demonstrate how the LLM performs on missing modalities without any prompts, we establish a \textbf{baseline} by employing a linear classifier head on the same LLM and randomly selecting feature tokens for its input. Results show that the performance significantly decreased by 2.62\% on pathology only, 1.55\% on genomics only and 2.05\% on complete test data. This indicates the importance of appropriate prompts for LLMs. (2) When supplementing modality-specific information for missing modalities, the performance (+UD) consistently improves on all datasets and increases by around +1\% on various inference settings, which validates the effectiveness of compensating for modality-specific knowledge of the missing modality. (3) When reusing UniPro to select relevant tokens (+US), the performance increased by about +1\% overall, which benefits from the assistance of US in learning modality-common knowledge from WSIs and genes with high redundancy.

\begin{figure*}
\vspace{-0.2cm}
    \centering
    \includegraphics[scale=0.26]{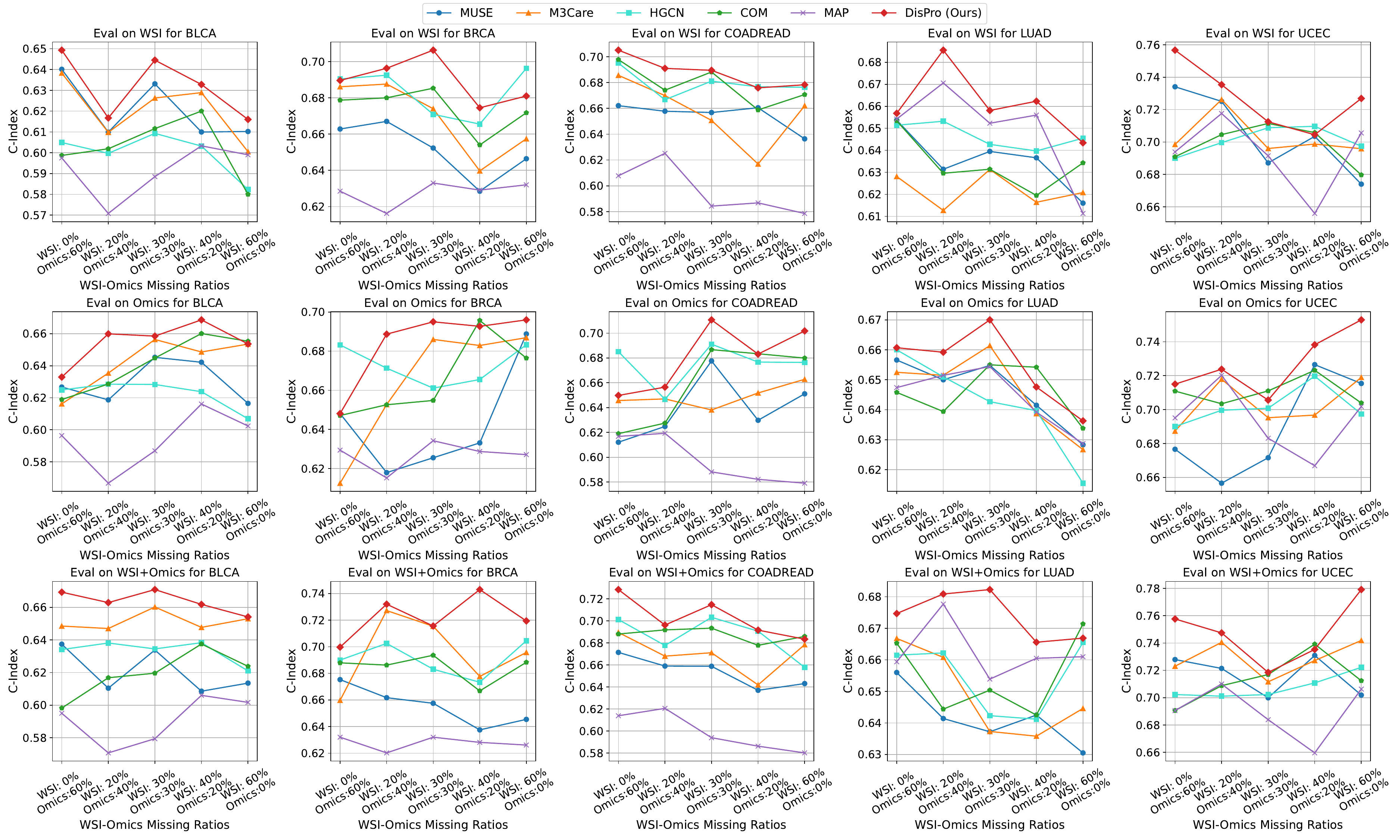}
    \caption{Performance on various combinations under 60\% training missing rate for different test scenarios.}
    \vspace{-0.2cm}
    \label{fig:missing_curve}
\end{figure*}

\noindent
\textbf{Various Missing Scenarios}. To illustrate how incomplete multimodal approaches handle varying missing settings, we display different combinations of training missing rates for each inference scenario, as shown in Fig.~\ref{fig:missing_curve}. Results showcase that DisPro achieved considerably consistent performance superiority in various missing settings over other incomplete multimodal approaches by a clear margin.

\section{Conclusion}
\label{sec:conclude}
This work proposed a Distilled Prompt Learning (DisPro) framework to unlock the potential of LLMs for incomplete multimodal learning by two-stage prompting, in which Unimodal Prompting (UniPro) and Multimodal Prompting (MultiPro) were proposed for supplementing modality-specific and modality-common knowledge for missing modalities. In the first stage, UniPro extended CoOp into the MIL setting and distilled knowledge distribution to a set of learnable prompts for each modality, which prepped for compensation of modality-specific knowledge. In the second stage, considering available modalities as prompts, we leveraged an LLM to impute common knowledge for missing modalities. To inject the specific knowledge of missing modalities, the distilled prompts in UniPro are used to supervise the learning of imputing representations for missing modalities, leading to more comprehensive knowledge compensation in incomplete multimodal learning. Extensive experiments on five cancer benchmark datasets demonstrated the effectiveness of the proposed method over state-of-the-art approaches.

\section*{Acknowledgement}
This work was supported by the National Natural Science Foundation of China (No. 62202403), Hong Kong Innovation and Technology Commission (Project No. MHP/002/22 and ITCPD/17-9),  and Research Grants Council of the Hong Kong Special Administrative Region, China (Project No. R6003-22 and C4024-22GF).
{
    \small
    \bibliographystyle{ieeenat_fullname}
    \bibliography{main}
}

\clearpage
\setcounter{page}{1}
\maketitlesupplementary
In this supplementary materials, we will include details of datasets and implementation, as well as more experimental results.

\section{Datasets}
The datasets involved in this paper are sourced from The Cancer Genome Atlas (TCGA) including Bladder Urothelial Carcinoma (BLCA), Breast Invasive Carcinoma (BRCA), Colon and Rectum Adenocarcinoma (COADREAD), Lung Adenocarcinoma (LUAD) and Uterine Corpus Endometrial Carcinoma (UCEC). We used all paired 20$\times$ WSIs and RNA-Seq data to evaluate overall survival (OS)~\cite{xu2023multimodal,zhou2023cross}. After removing the absent genes, all genes are grouped into 330 pathways provided by the previous work~\cite{jaume2024modeling}. The detailed list of genes will be present in the code repository upon acceptance.
\begin{table*}[htbp]
\small
  \centering
    \begin{tabular}{c|cc|ccc|ccccc|c}
    \toprule
    \multirow{2}[4]{*}{Variants} & \multicolumn{2}{c|}{Test} & \multicolumn{3}{c|}{Modules} & \multicolumn{1}{c}{\multirow{2}[4]{*}{\makecell[c]{BLCA\\(372)}}} & \multicolumn{1}{c}{\multirow{2}[4]{*}{\makecell[c]{BRCA\\(1007)}}} & \multicolumn{1}{c}{\multirow{2}[4]{*}{\makecell[c]{COADREAD\\(533)}}} & \multicolumn{1}{c}{\multirow{2}[4]{*}{\makecell[c]{LUAD\\(443)}}} & \multicolumn{1}{c|}{\multirow{2}[4]{*}{\makecell[c]{UCEC\\(478)}}} & \multirow{2}[4]{*}{Avg} \\
\cmidrule{2-6}          & P     & G     & Self  & Cross & Uni   &       &       &       &       &       &  \\
    \midrule
    w/o US & $\bullet$ & $\circ$ &       &       &       & 0.6213±0.010 & 0.6829±0.008 & 0.6740±0.022 & 0.6327±0.018 & 0.7152±0.011 & 0.6652 \\
     + Self & $\bullet$ & $\circ$ & \checkmark &       &       & 0.6095±0.007 & 0.6742±0.014 & 0.6748±0.016 & 0.6512±0.007 & 0.7187±0.011 & 0.6657 \\
     ++ Cross & $\bullet$ & $\circ$ & \checkmark & \checkmark &       & 0.6208±0.010 & 0.6860±0.006 & 0.6800±0.016 & 0.6548±0.009 & \textbf{0.7291±0.010} & 0.6741 \\
     ++ Uni & $\bullet$ & $\circ$ & \checkmark &       & \checkmark & 0.6214±0.008 & 0.6814±0.008 & 0.6795±0.025 & 0.6512±0.009 & 0.7196±0.016 & 0.6706 \\
     \rowcolor{red!10}
    DisPro (full) & $\bullet$ & $\circ$ & \checkmark & \checkmark & \checkmark & \textbf{0.6319±0.014} & \textbf{0.6895±0.011} & \textbf{0.6880±0.010} & \textbf{0.6612±0.014} & 0.7272±0.018 & \textbf{0.6796} \\
    \midrule
    w/o US & $\circ$ & $\bullet$ &       &       &       & 0.6473±0.017 & 0.6836±0.014 & 0.6707±0.022 & 0.6420±0.020 & 0.7226±0.013 & 0.6732 \\
     + Self & $\circ$ & $\bullet$ & \checkmark &       &       & 0.6514±0.013 & 0.6763±0.018 & 0.6662±0.026 & 0.6468±0.017 & 0.7191±0.020 & 0.6720 \\
     ++ Cross & $\circ$ & $\bullet$ & \checkmark & \checkmark &       & 0.6532±0.020 & 0.6825±0.011 & 0.6754±0.032 & 0.6484±0.015 & 0.7237±0.023 & 0.6766 \\
     ++ Uni & $\circ$ & $\bullet$ & \checkmark &       & \checkmark & 0.6498±0.014 & 0.6810±0.009 & 0.6637±0.026 & 0.6487±0.023 & 0.7199±0.013 & 0.6726 \\
    \rowcolor{red!10}
    DisPro (full) & $\circ$ & $\bullet$ & \checkmark & \checkmark & \checkmark & \textbf{0.6547±0.012} & \textbf{0.6841±0.018} & \textbf{0.6804±0.024} & \textbf{0.6548±0.012} & \textbf{0.7271±0.017} & \textbf{0.6802} \\
    \midrule
    w/o US & $\bullet$ & $\bullet$ &       &       &       & 0.6585±0.007 & 0.7194±0.013 & 0.6963±0.009 & 0.6635±0.014 & 0.7250±0.012 & 0.6926 \\
     + Self & $\bullet$ & $\bullet$ & \checkmark &       &       & 0.6499±0.009 & 0.7091±0.006 & 0.6959±0.014 & 0.6590±0.009 & 0.7315±0.009 & 0.6891 \\
     ++ Cross & $\bullet$ & $\bullet$ & \checkmark & \checkmark &       & 0.6608±0.007 & 0.7190±0.013 & 0.7015±0.025 & 0.6646±0.020 & 0.7324±0.009 & 0.6957 \\
     ++ Uni & $\bullet$ & $\bullet$ & \checkmark &       & \checkmark & 0.6582±0.003 & 0.7183±0.012 & 0.6992±0.015 & 0.6659±0.017 & 0.7263±0.019 & 0.6936 \\
    \rowcolor{red!10}
    DisPro (full) & $\bullet$ & $\bullet$ & \checkmark & \checkmark & \checkmark & \textbf{0.6638±0.006} & \textbf{0.7219±0.015} & \textbf{0.7029±0.016} & \textbf{0.6741±0.007} & \textbf{0.7476±0.020} & \textbf{0.7021} \\
    \bottomrule
    \end{tabular}%
    \caption{\textbf{Ablation Study (C-Index) on UniPro Scoring (US) under 60\% training missing rate}. `Self' refers to the \textit{Self-Scoring} module in US. `Uni' indicates considering the textual class representation of the query modality when calculating scores in Eq. 8, while `Cross' indicates considering the textual class representation of non-query modalities.}
  \label{tab:abla_us}%
\end{table*}%

\begin{figure*}
    \centering
    \vspace{-0.3cm}
    \includegraphics[width=1.0\linewidth]{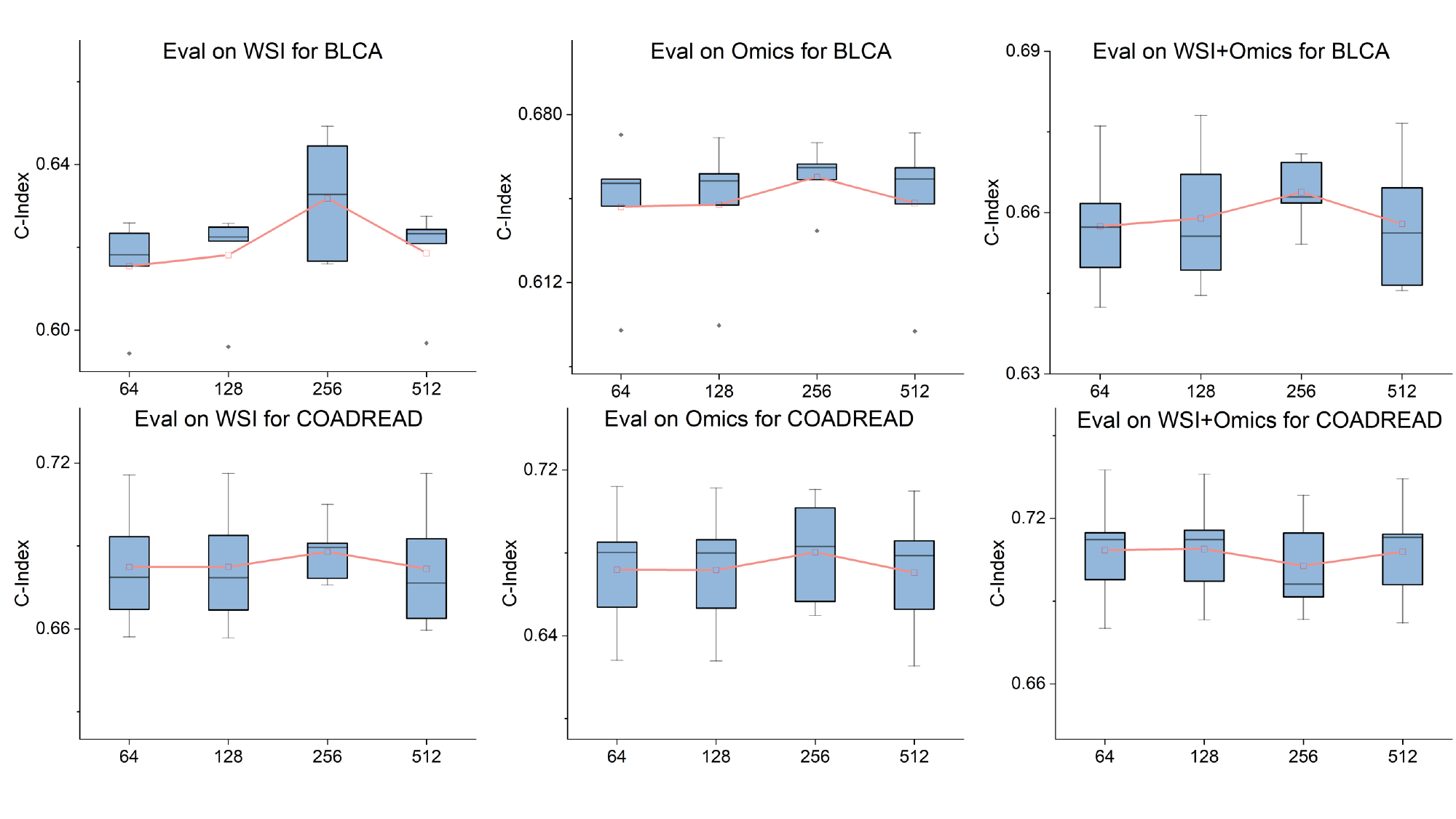}
    \caption{Performance on various K of Top-K MaxPooling in UniPro Scoring.}
    \vspace{-0.3cm}
    \label{fig: exp_topk}
\end{figure*}
\section{Implementation}
\subsection{Model}
\textbf{Stage 1 - UniPro.} In terms of UniPro for Pathology, a pathology foundation model, UNI~\cite{chen2024towards}, was used as the patch encoder to extract a 1024-d feature for each patch image. A linear layer followed by ReLU~\cite{brownlee2019gentle} was employed as the adapter to project each patch feature into a 768-d feature. Similarly, SNN~\cite{klambauer2017self} was used to encode each pathway into 768-d feature, leading to a sequence of tokens with the shape of (330, 768).

To learn the morphological descriptions in WSIs for every risk band of survival outcome, we set up 8 groups of prompts for the textual branch corresponding to the category of the pair of survival time (4 bins of time intervals, i.e., $\boldsymbol{I}_t=4$) and censorship status (0 or 1). For every prompt of each category, three parts include learnable context tokens, prefix tokens and classname tokens. The sequence length of learnable context tokens is set as 255 (i.e., $k=255$ in Eq. 3 and 4) and the embedding dimension of each token is 768, leading to a set of learnable context tokens with the shape of (8, 255, 768) concatenated into the [CLS] token of textual input of UniPro-P along the dimension of the sequence length. 

Then, the pretrained tokenizer of BioBERT-v1.2~\cite{lee2020biobert} was used to tokenize the prefix words of ``\textit{This is a pathology slide image from the patient with overall survival of}'' into prefix tokens with the embedding dimension of 768, which are shared across different classes. Prefix tokens are put right after the context tokens along the sequence dimension. 

The classname tokens were also obtained by embedding a group of classnames into 768-d tokens using the same tokenizer, including 1) \textit{high risk, dead}, 2) \textit{mid-high risk, dead}, 3) \textit{mid-low risk, dead}, 4) \textit{low risk, dead}, 5) \textit{short observation, alive}, 6) \textit{mid-short observation, alive}, 7) \textit{mid-long observation, alive} and 8) \textit{long observation, alive}. These classname tokens were put at the end of the textual input after prefix tokens, leading into the final textual input $\boldsymbol{T}_p$. By forwarding the constructed textual input $\boldsymbol{T}_p$ into a LLM, BioBERT-v1.2~\cite{lee2020biobert}, the [CLS] token of each class prompt in the output of the LLM was used as the representation $\boldsymbol{t}_p^{(j)}$ of a class $j$, which would be substituted into Eq. 5 for top-K MaxPooing (K=256).

Similarly, to capture the expression patterns in RNA-Seq data for every risk band, we set up prompts following the aforementioned steps. In particular, the sequence length of genomic context tokens is 256, and prefix words become ``\textit{These are gene expression profiles from the patient with overall survival of}''. Other settings are the same as UniPro-P.

\noindent
\textbf{Stage 2 - MultiPro}. In this stage, we used the same LLM, BioBERT-v1.2~\cite{lee2020biobert} to encode the multimodal input, which consisted of 3 parts: a [CLS] token, a sequence of pathology feature tokens and a sequence of genomic feature tokens. The maximum length of input for BioBERT-v1.2 is 512, and thus we set the lengths of pathology and genomics by 255 and 256 (i.e., $K_p=255$ and $K_g=256$), respectively. The dimension of each token is 768 as well. Additionally, $K=256$ is set in Top-K MaxPooling during inference of UniPro Scoring, unless otherwise specified. The coefficient factors $\alpha_1$ and $\alpha_2$ are simply set by 1.0 and 1.0, respectively.
\subsection{Training}
The setting for missing modalities is introduced in the main text. Here we present the details of the training procedure.

Following the previous setting~\cite{xu2023multimodal,zhou2023cross}, we adopted Adam optimizer with the initial learning rate of $2\times 10^{-4}$ and weight decay of $1\times 10^{-5}$, unless otherwise specified. Due to the large size and varying length of WSIs, the batch size is 1 following the common setting~\cite{chen2021multimodal,xu2023multimodal}. All experiments are trained for 30 epochs by default to ensure the convergence of every model. Particularly, for prompt-based methods that employed BioBERT, the initial learning rate of genomic backbone becomes $1\times 10^{-5}$, and these models are trained for 50 epochs to guarantee the full convergence of these models. For a fair comparison, all models are ensured to have fully converged.

\section{More Ablation Studies}
\subsection{Various Types of Tokens in UniPro Scoring}
In this section, we investigate the roles of various tokens in UniPro Scoring, and results are shown in Tab.~\ref{tab:abla_us}. Take the query modality for the UniPro Scoring module to be pathology as an example. The query tokens are a bag of features tokens $\left\{\mathbf{p}_n^i\right\}_{i=1}^{M_p}$. Then, `Self' refers to the scores $\mathbf{a}_{n, p}$ computed by attention layers. `Cross' suggests that the similarity between query features tokens and genomic textual class representation $\boldsymbol{t}_g$ to get the scores $\mathbf{s}_{n, g}$ in Eq. 8. Similarly, `Uni' suggests that the similarity between query features tokens and pathological textual class representation $\boldsymbol{t}_p$ to get the scores $\mathbf{s}_{n, p}$.

We observed that 1) when only incorporating Self-Scoring module, the performance become worse than the variant without US. The possible reason could be the information included in incomplete data is not enough for training a robust grader from scratch. 2) when additionally introducing either `Cross' (++ Cross) or `Uni' (++ Uni) scoring, the performance consistently surpasses the variant without US. This indicates the distilled knowledge can bring extra performance gains. In particularly, `Cross' contributes to performance increases more significantly than `Uni', which could be attributed to the assistance of `Cross' in compensating for the modality-common knowledge. 3) The full version of DisPro achieves the best performance, suggesting the modal can benefit from their collaboration.
\subsection{Top-K MaxPooling in UniPro Scoring}
In this part, we explore the effect of the selection of K on DisPro. We set up a series of K including 64, 128, 256 and 512. Note that in other experiments, K is always 256. Results are shown in Fig.~\ref{fig: exp_topk}. In most cases, as K increases, the performance gradually rises until it reaches a peak, after which further increases in K will not result in significant performance gains. Therefore, considering the trade-off between performance and inference speed, we take K=256 as our default setting.

\section{Visualization Interpretation}
\begin{figure*}
\centering
\includegraphics[scale=1.0]{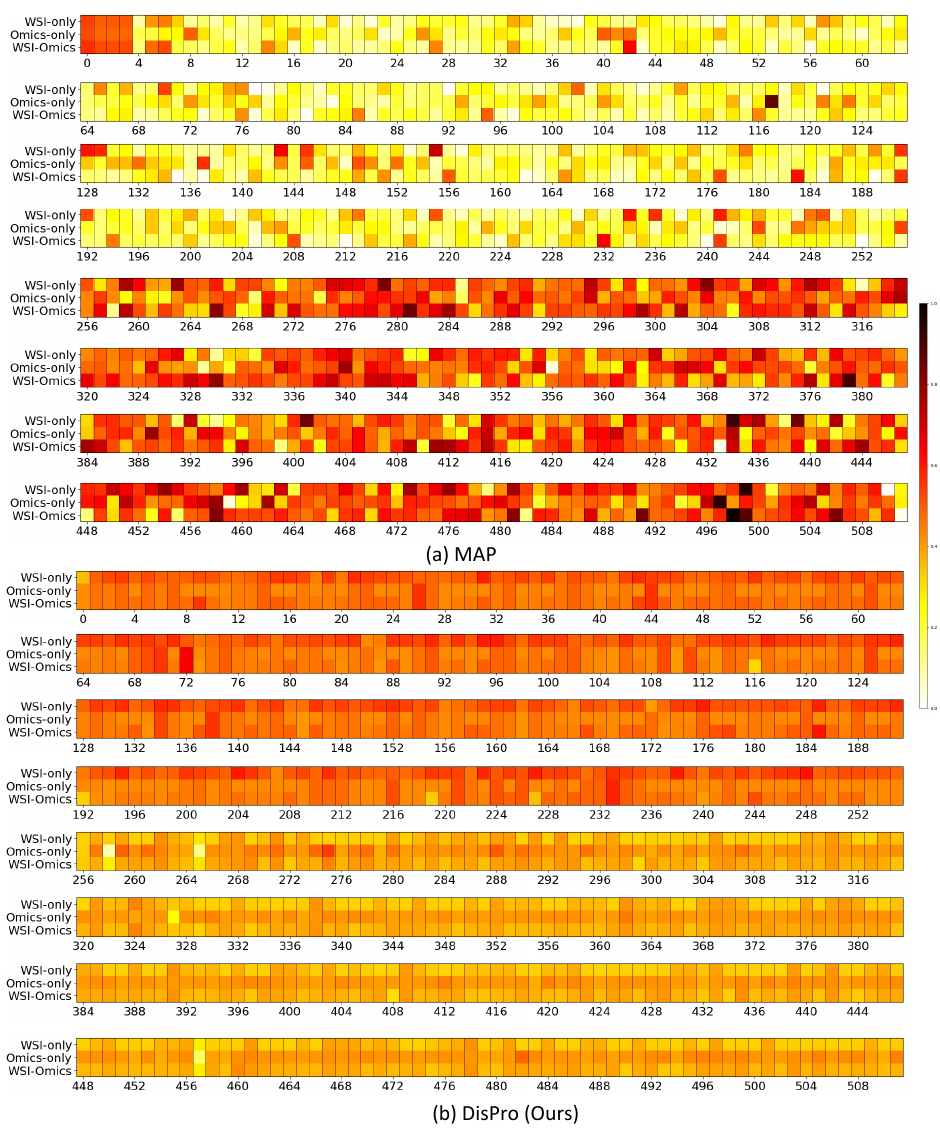}
\caption{The visualization of attention signals for each token in LLM used in (a) MAP and (b) DisPro (Ours).}
\label{fig:visualize}
\end{figure*}

To intuitively validate if DisPro is robust to missing modalities, we visualize the attention signals of each token (512 in total for BioBERT) in LLM under the situations of missing modalities (WSI-only or Omics-only) and complete modality (WSI-Omics), and compare the prompt-based model, MAP~\cite{lee2023multimodal}. We feed different combinations of missing and complete modalities of the same sample to the model and observe the differences of attention signals among them. If the model fed by incomplete data can predict similar attention signals to that of complete modality, the robustness to missing modalities has been learned by the model. The results are shown in Fig.~\ref{fig:visualize}, where we can see that attention signals for missing modalities in DisPro are more aligned with those of complete data, whereas MAP’s predictions are chaotic across various modality combinations. This indicates that DisPro is more robust to missing modalities.

\end{document}